\documentclass[10pt,twocolumn,letterpaper]{article}
\usepackage{amsmath}
\usepackage{mathptmx}
\usepackage{wacv}
\usepackage{url}
\usepackage{times}
\usepackage{graphicx}
\usepackage{amssymb}
\usepackage{subcaption}
\usepackage{fancyref}
\usepackage{color}

\usepackage[pagebackref=true,breaklinks=true,letterpaper=true,colorlinks,bookmarks=false]{hyperref}

\wacvfinalcopy 

\def\wacvPaperID{183} 

\begin{document}

\title{Egocentric Height Estimation}

\author{Jessie Finocchiaro \\
Florida Southern College\\
{\tt\small jfinocchiaro@mocs.flsouthern.edu}
\and
Aisha Urooj Khan \\
University of Central Florida\\
{\tt\small aisha.urooj@knights.ucf.edu}
\and
Ali Borji \\
University of Central Florida\\
{\tt\small aborji@crcv.ucf.edu}
}

\maketitle
\ifwacvfinal\thispagestyle{empty}\fi

\begin{abstract}
Egocentric, or first-person vision which became popular in recent years with an emerge in wearable technology, is different than exocentric (third-person) vision in some distinguishable ways, one of which being that the camera wearer is generally not visible in the video frames. Recent work has been done on action and object recognition in egocentric videos, as well as work on biometric extraction from first-person videos. Height estimation can be a useful feature for both soft-biometrics and object tracking. Here, we propose a method of estimating the height of an egocentric camera without any calibration or reference points. We used both traditional computer vision approaches and deep learning in order to determine the visual cues that results in best height estimation. Here, we introduce a framework inspired by two stream networks comprising of two Convolutional Neural Networks, one based on spatial information, and one based on information given by optical flow in a frame.  Given an egocentric video as an input to the framework, our model yields a height estimate as an output.  We also incorporate late fusion to learn a combination of temporal and spatial cues.  Comparing our model with other methods we used as baselines, we achieve height estimates for videos with a Mean Average Error of  14.04 cm over a range of 103 cm of data, and classification accuracy for relative height (tall, medium or short) up to 93.75\% where chance level is 33\%.

\end{abstract}

\section{Introduction}

\noindent Growing usage of wearable devices in recent years is made possible with products like Google Glass \cite{googleglass}, GoPro \cite{gopro}, and Narrative Clip \cite{narclip} becoming increasingly affordable. These wearable cameras are generating huge amount of data which requires automatic analysis, so that useful applications are developed for these devices \cite{reyesdid}. To meet these needs, egocentric video analysis is now one of the emerging domains in computer vision.

\noindent Most of the work egocentric vision encompasses is related to camera wearer's activities and behavior. Nonetheless, we can also tell much about the identity of camera wearer \cite{Peleg2}, which inspired us to be curious about the questions related to appearance of the camera wearer. We can ask many questions looking at a first person video. Some questions are: \textit{Can we estimate photographer's height, gender, mental disorders, movement disorders, etcetera by just looking at the first person perspective video?} Here, we focus on height estimation of camera wearer as it has potential for surveillance and biometric applications.  In particular, height estimation can be helpful to track the same person in static cameras which may be possibly helpful for surveillance purposes. Identifying the person in a static camera by looking at the egocentric view is an interesting question. \cite{ardeshiregocentric} have formulated this problem as a graph matching problem and used unary and pairwise features to acheive significant improvement. Kisp{\'a}l \cite{humanhe} used camera calibration for estimating height of humans present in a video. Similar work has been done by \cite{tpv}, where a single image is used for estimating body height. \cite{surveillance} used surveillance camera footage for height estimation and used single view metrology method which they referred as SVM to estimate height. \cite{singlecamerahe} did similar work with single camera view. However, our problem is unique in the sense that we attempt to estimate height in egocentric videos where the person is invisible, also we do not use any camera calibration for the purpose.
To the best of our knowledge, no work has been done previously to address this problem in egocentric videos.

\noindent As we find no previous work in this line of research, no publicly available egocentric datasets could be useful for our experiments. Thus, we collected our own dataset of 60 videos featuring 10 participants in two background settings; static where no moving objects are in the background, and dynamic where people are walking around the camera wearer. These videos are collected from three different heights - the waist, chest and head for each person, to simulate three different people: short, average and tall. These videos have varying length range from 23 \textit{seconds} to 1 \textit{minute} 16  \textit{seconds} depending on how fast or slow a person walks when allowed to move with her normal pace.

\noindent To start with, our first approach for estimating height was to train a Support Vector Classifier (SVC) to determine if we could accurately determine which camera height, or ``person," an image was taken from. After, we discretized the camera heights into smaller bins to see the accuracy we could attain with smaller height ranges. we tested our SVC on different extracted features from video frames such as HOG, SIFT, GIST and features from a pre-trained convolutional neural network. 
	
\noindent To estimate wearer's height continuously (in recorded video), we modeled a CNN based on Peleg's in \cite{Peleg}. While their network is used as an action recognition classifier, we re-implemented (in Python) and transformed their network output to be regression-based, and used it to estimate height in shorter clips of video, instead of a 14-way classifier output like their work. We then adjusted the network further to consider the spatial information in the video. In their network, optical flow frames are concatenated over the temporal axis and 3D convolution is used in their video. Instead of taking sparse optical flow as an input, the spatial network was adjusted to take grayscale images as input. We used Exponential Linear Units (ELU) as activation function as well as making minor adjustments to the stride and kernel size of one of the Convolutional layers since the grayscale images are one layer instead of two.
	
\noindent Last, we combined the output of the temporal and spatial CNNs in a two-stream CNN to see if the combined networks could improve our results, as in \cite{twostream}.  We performed fusion at two different layers of the network to see if results improved with earlier fusion.

\begin{figure*}
\begin{subfigure}{.33\textwidth}
  \centering
  \includegraphics[width=\textwidth]{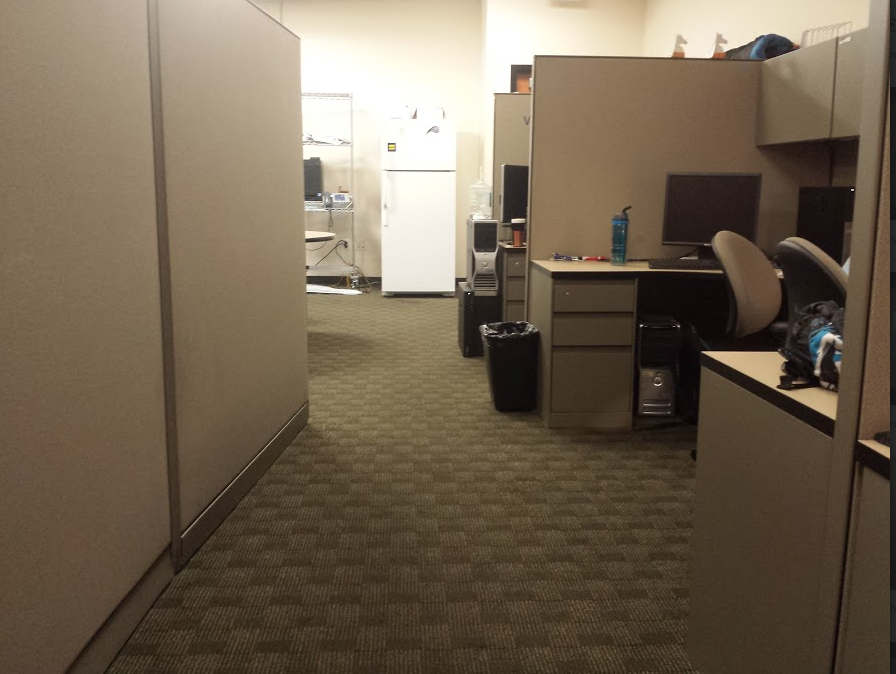}
  \caption{Image taken from waist-mounted camera}
  \label{fig:waistcam}
  \end{subfigure}
  \begin{subfigure}{.33\textwidth}
  \centering
  \includegraphics[width=\textwidth]{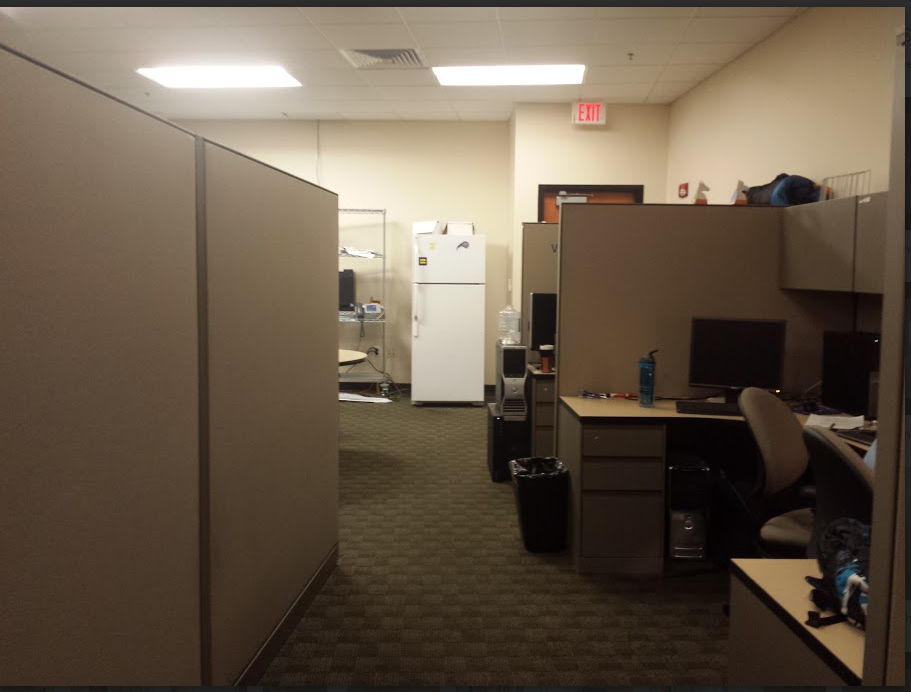}
  \caption{Image taken from chest-mounted camera}
  \label{fig:chestcam}
  \end{subfigure}%
\begin{subfigure}{.33\textwidth}
  \centering
  \includegraphics[width=\textwidth]{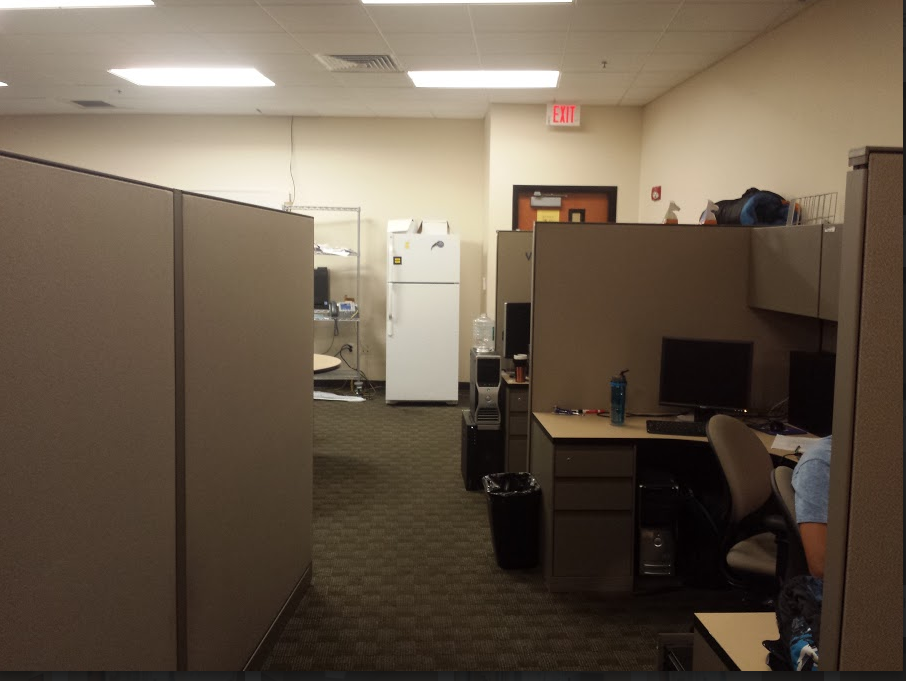}
  \caption{Image taken from head-mounted camera}
  \label{fig:headcam}
  \end{subfigure}%

  \caption{Three pictures taken in the same spot from different heights.}
   \label{fig:3samples}
\end{figure*}

\noindent In what follows, we present a brief survey of related work in Section 2.  We describe our EgoHeights dataset in Section 3.  We took two approaches to estimate approximate height of the camera, and we describe both in Section 4.  Results are shown and described in Section 5.  We expand our network training and test robustness to investigate overfitting in Section 6 and conclude our work in Section 7.
	
\section{Related Works}
\noindent With increasing popularity of egocentric vision in recent years, much research work is being focused to solve computer vision problems with the perspective of first person. Egocentric or First Person Vision problems are unlike classic computer vision problems since the person whose actions are being recorded is not captured. Egocentric vision poses unique challenges like non-static cameras, unusual view points, motion blur \cite{Bambach_2015_ICCV}, variations in illumination with the varying positions of camera wearer, real time video analysis requirements, etcetera \cite{betancourt2015evolution}. Tan \textit{et al.} \cite{fpvvstpv} demonstrate that challenges posed by egocentric vision can be handled in a more efficient manner if analyzed differently than exocentric. Much work has been done to address classic computer vision problems, now with egocentric perspective such as objects understanding \cite{Lee_2014_CVPR_Workshops}  \cite{Zhou_2016_CVPR}, object detection \cite{reyesdid} \cite{DBLP:journals/corr/RogezSKMR14} \cite{Bambach_2015_ICCV}, object tracking \cite{DBLP:journals/corr/WangCF16} \cite{DBLP:journals/corr/AghaeiDR15}, and activity recognition \cite{Peleg} \cite{delving} \cite{dogcentric} \cite{posneg}. 

\noindent Recently, Zhou \textit{et al.} \cite{Zhou_2016_CVPR} proposed a cascaded interactional targeting deep neural network for egocentric action recognition. In \cite{social}, saliency cues are exploited to predict objects of attention in egocentric videos. Another line of work is related to summarizing hours long videos recorded by camera wearers and predicting important objects, people and interesting events in life-logging data \cite{Xu_2015_CVPR} \cite{lee2015predicting} \cite{Lu_2013_CVPR} \cite{lee2012discovering}. Albeit the camera wearer is not captured and apparently seems to preserve his anonymity, nevertheless, much can be told about the person \cite{Peleg2}. Peleg \textit{et al.} \cite{Peleg2} used camera motion cue to demonstrate how a person's identity can be revealed. Estimating height of objects or people in a video with a calibrated camera \cite{tpv} \cite{humanhe} \cite{surveillance} \cite{singlecamerahe} is also reported.  In \cite{3dpose}, dynamic motion signatures and static scene structures proved to be significant cues for pose estimation of the camera wearer when most of the significant body parts are invisible.  Our problem, however, is unique since the person whose height we are trying to estimate is not in the image frame and using an uncalibrated camera further complicates the problem. 
\begin{figure}[b!]
	\includegraphics[width=\linewidth]{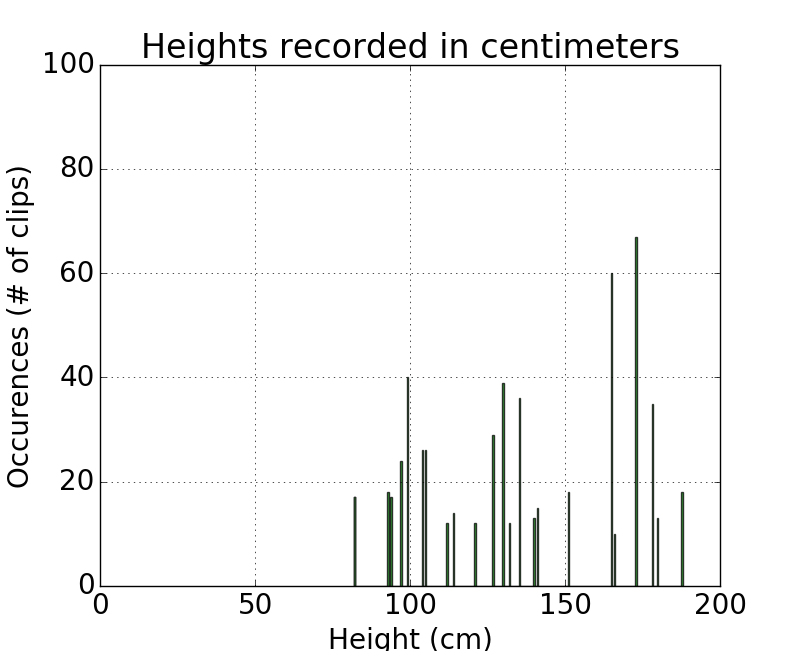}
	\caption{Distribution of video segments by height in centimeters.  Mean is 136.14 cm and standard deviation is 31.525 cm}
	\label{fig:heighthist}
\end{figure}
\section{EgoHeights Dataset}
\noindent We collected our own dataset for this work because there was no way of obtaining the knowledge to annotate a previously published dataset fitting our needs. To collect this dataset, participants were asked to walk down a hallway with a mounted camera on their body. We collected data in two different background settings; \textit{Static}: where no objects or people were moving in the background, and \textit{Dynamic}: people are walking around the camera wearer. For the static setting, participants were asked to make a smooth walking motion, whereas, for the dynamic setting, participants were allowed to move with their normal pace. 
The data was collected for 10 participants with three different camera positions (\textit{head, chest and waist}) for both static and dynamic background resulting in total 10 x 3 x 2 = 60 videos. Gender ratio for our participants was 7 males and 3 females. Each participant wore Samsung galaxy S4 with straps for head, chest or waist which recorded 270 x 480 pixels video at 30 \textit{fps}.Video lengths range from 23 \textit{seconds} to 1 \textit{minute} 18 \textit{seconds}.
 In total, we collected total 64,830 \textit{ frames} (33780 \textit{ frames} for static environment and 31050 \textit{ frames} for dynamic environment). For ground truth, each of the videos has been given a relative label (\textit{short, medium or tall}), and an absolute label (the absolute height of the camera). Camera height was measured in centimeters with smallest height of 85 \textit{cms} and largest height of 188 \textit{cms}. In Figure \ref{fig:3samples}, we demonstrate the difference in image view with different camera heights when three images are taken in the same spot.
 Figure \ref{fig:heighthist} shows the distribution of heights from which the video was recorded in relation to the length of video.  
It is important to note that we estimated the height of the camera, not the height of the person regardless of camera location. 

\section{Approach}
\subsection{Support Vector Classifier (SVC)}
\begin{figure}[b!]
	\includegraphics[width=\linewidth]{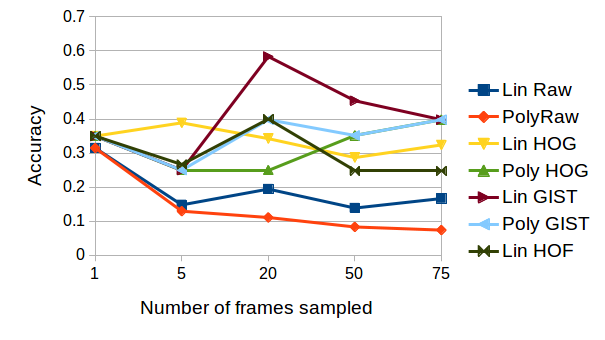}
	\caption{Accuracy of different image descriptors and SVM kernels with random samples of frames from each training video.}
	\label{fig:tempframes}
\end{figure}

\noindent We trained a SVC to classify frames for three different ranges of height. We did this to extract information about what factors contribute to egocentric height estimation.  Different image descriptors were used to train the SVC as we wanted to see what properties contributed to successful height estimation.  To train the SVC, we used Linear and Polynomial kernels on different image descriptors and performed 3-fold cross-validation on the dataset.  Different image features and weights from a pre-trained neural network \cite{AlexNet} were extracted and vectorized to train the classifier for frames in the video.  We also trained classifiers on the Raw Image Vectors as a baseline comparison.  Leave-One-Out testing was performed to see the results for each person's video with 3 bins, 5 bins, and 11 bins.  
Additionally, we performed this testing using subsets of each video to train.  We selected 1, 5, 20, 50, and 75 frames at random from each training video and tested the last video using different kernels and feature descriptors.  Results are seen in Figure \ref{fig:tempframes}. Results initially are high because of overfitting, but after an initial decrease, accuracy raises again as the SVM fits the kernel to the data. 

\subsubsection{3 bin classifier}
\noindent The first bins we used to classify video was the location of the camera with respect to the recorder's body (ie. waist, chest, or head.)  Feature extractors such as Histograms of Oriented Gradient (HOG), SIFT \cite{sift}, Raw Image Vectors, Histograms of Oriented Optical Flow, GIST features, and various layers of pre-trained AlexNet \cite{AlexNet} were used to train and test every 15th frame in a video.  The results were validated with 3-fold cross-validation.  Histograms of Oriented Gradient yielded the best results, well above chance for all three classifiers.  After HOG, AlexNet layers consistently yielded the best results.  Because of the success of pre-trained Alexnet, we compared an SVM based on our temporal and spatial networks as well as a classification based version of our network, with the last activation layer changed to be a softmax, both of which outperformed HOG as seen in Figures \ref{fig:sfig4} to \ref{fig:sfig6}.  Our classifying network yields classification accuracies up to 93.75\%.  Because of this, we assume we can proceed using our network to estimate continuous height.  In order to incorporate noise, we repeated our method with the videos recorded with dynamic backgrounds in Figure \ref{fig:svmresults}.    While some accuracy is lost training on a static background and testing on dynamic, we still report classification accuracies above chance, and HOG is still the most successful image descriptor, with a 3 bin accuracy of 55.25\%, compared to the chance classification of 33\% accuracy.

\subsubsection{5 and 11 bin classifiers}
\noindent Due to our small dataset, the test subjects whose results had the lowest accuracies are those heights which happened to be the extremes in our dataset.  Because somebody who might be very tall might record their chest camera at a height close to the height of the head camera of a short person, we proceeded by training the SVC on smaller bins by centimeter height instead of camera location.  
 
\begin{figure*}
\begin{subfigure}{.33\textwidth}
  \centering
  \includegraphics[width=\textwidth]{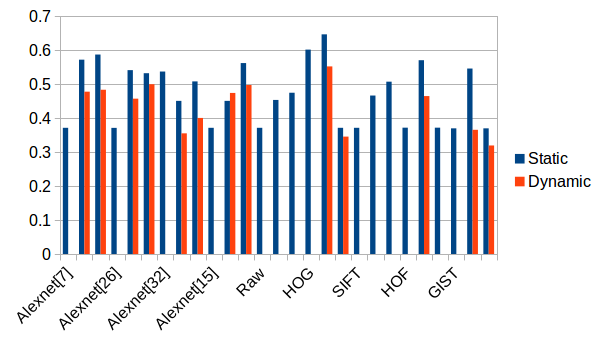}
  \caption{3 bins}
  \label{fig:sfig1}
\end{subfigure}%
\begin{subfigure}{.33\textwidth}
  \centering
  \includegraphics[width=\textwidth]{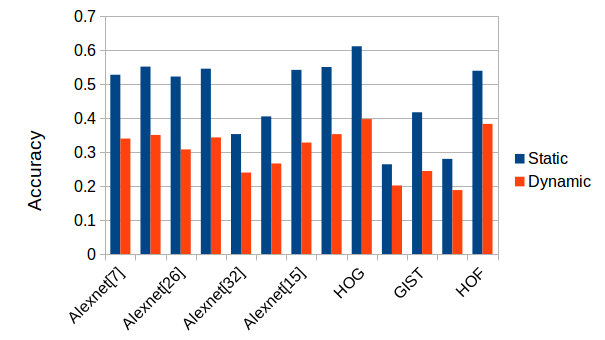}
  \caption{5 bins}
  \label{fig:sfig2}
\end{subfigure}%
\begin{subfigure}{.33\linewidth}
  \centering
  \includegraphics[width=\linewidth]{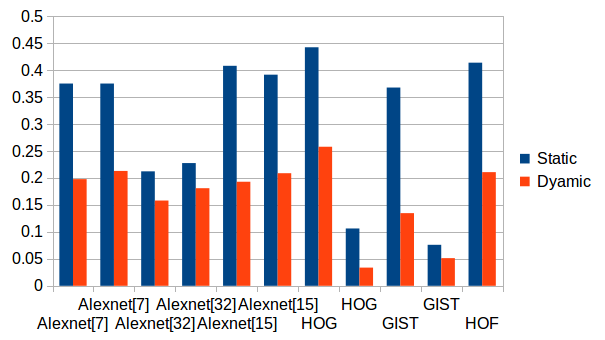}
  \caption{11 bins}
  \label{fig:sfig3}
\end{subfigure}

\begin{subfigure}{.33\textwidth}
  \centering
  \includegraphics[width=\textwidth]{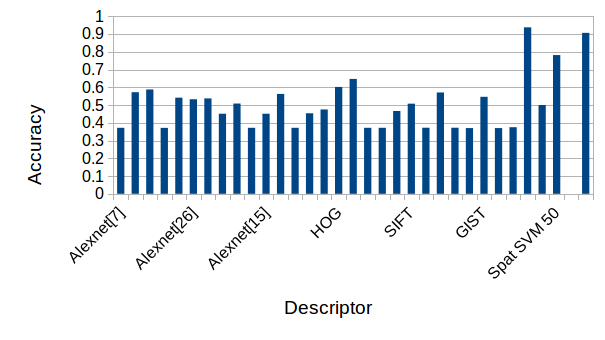}
  \caption{3 bins on clean frames}
  \label{fig:sfig4}
\end{subfigure}%
\begin{subfigure}{.33\textwidth}
  \centering
  \includegraphics[width=\textwidth]{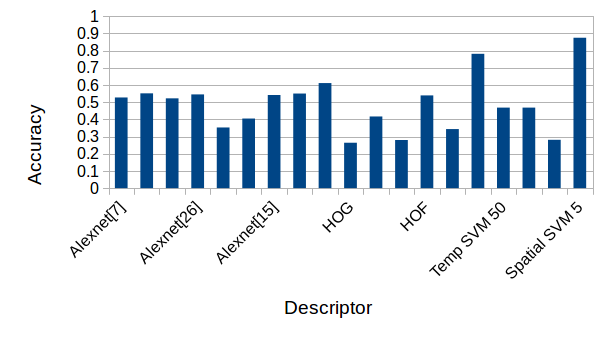}
  \caption{5 bins on clean frames}
  \label{fig:sfig5}
\end{subfigure}%
\begin{subfigure}{.33\linewidth}
  \centering
  \includegraphics[width=\linewidth]{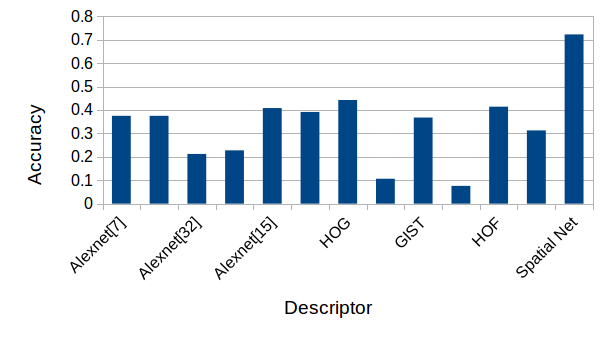}
  \caption{11 bins on clean frames}
  \label{fig:sfig6}
\end{subfigure}
\caption{After our spatial network's output, Linear HOG consistently yielded the best results, followed by different layers of AlexNet and HOF.  Linear kernel then polynomial shown for each descriptor.}
\label{fig:svmresults}
\end{figure*}

\noindent To start, we split data into 25 cm bins, yielding 5 bins with the range of our data.  With a chance accuracy of 20\%, results are shown in \ref{fig:sfig2} and \ref{fig:sfig5} for the different feature descriptors.  HOG is still the most successful image descriptor with accuracy of 61.15\% on static images and 39.73\% on dynamic frames.  However, our network yields a classification accuracy of 78.15\%.
 
\noindent Similarly, we split the data into 10 cm bins, yielding 11 bins and repeated our training with a chance of 9\%.  Results are shown in \ref{fig:sfig3} and \ref{fig:sfig6}.  HOG yielded results with up to 44.25\% accuracy on static data and 25.80\% accuracy on dynamic, but our spatial network had accuracy of 72.23\%, well above the chance of 9\%.
 
\subsection{Convolutional Neural Networks (CNNs)}
\noindent Estimating height using a SVC has some drawbacks.  For example, heights that are on the ends of each bin range might actually be closer in height to data in another bin.  However, changing the output from classification to regression allows us to estimate height as a continuous measurement instead of in a discrete manner.
\begin{figure*}
	\includegraphics[width=.8\textwidth]{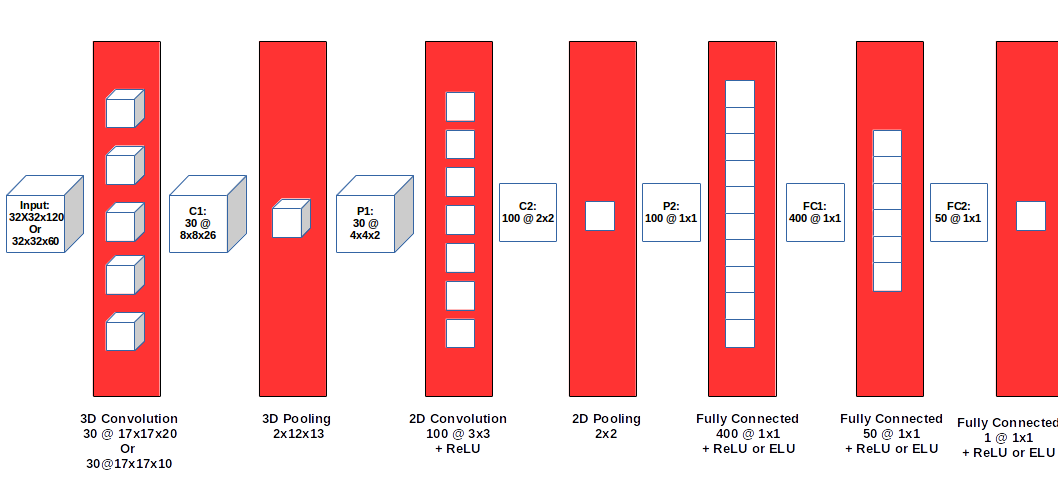}
    \centering	
	\caption{Model of our temporal network and spatial network.  The only difference is the first 3D Convolution layer.  Dimensions for the temporal network are given first, then dimensions for the spatial network.}
	\label{fig:indivnetwork}
\end{figure*}

\subsubsection{Temporal Network Architecture}

\noindent Our temporal network is based on Peleg et. al. in ~\cite{Peleg}.  However, their group used the network to classify actions in clips of egocentric video into one of 7 or 14 categories.  We modified the last layer of their network to have one output and used a linear activation function to estimate height.  We normalized input values between 0 and 1 so the network generally returned a number between 0 and 1.  We then converted the values back using the same scale to provide a height estimate.\\
 
\noindent Video segments were normalized to 15 frames per second.  A sparse 32x32 optical flow vector was extracted from the frames and the x and y values were temporally concatenated, then the frames over 4 seconds were concatenated in the same manner, making the input to the network 32x32x120 optical flow vectors.  Video collection started every 2 seconds so there was overlap in the videos.  A height estimate was taken for every 4 second clip, and the video's estimate was given by taking the average of estimates for each clip in a given video.
 
\noindent The network starts with a 3D convolutional layer with 30 kernels of size  17x17x20 with spatial stride of 2 and a temporal stride of 4, since flow vectors are concatenated on the temporal dimension.  3D pooling of size 2x2x13 is then applied with a stride of 2x2x13.  Output is size 4x4x2, and 2D Convolution is the next hidden layer with 100 kernels of size 3x3.  The output is 2x2, and max pooling is applied to make the output 1x1.  Fully connected layers of size 400, 50, and 1 are added to yield one output.  All activation functions are Rectified Linear Units (ReLU) except for the final neuron having a linear activation function.

\subsubsection{Spatial Network Architecture}
\noindent The architecture of our spatial network is very similar to that of the temporal network, but instead of taking optical flow as the input to the network, pixel intensities of grayscale images are used.  Because of that, the network input is 32x32x60.  Additionally, instead of using ReLU activation functions, we used an Exponential Linear Unit and Batch Normalization with our dataset. ~\cite{Peleg}

\subsubsection{Late Fusion of Networks}
\noindent We took our spatial and temporal networks and concatenated the weights of neurons at the last and second-to-last layers and concatenated them.  We then trained the network with a fully connected layer from either two neurons or 100 neurons to one neuron with linear activation.

\begin{figure*}
	\includegraphics[width=\textwidth]{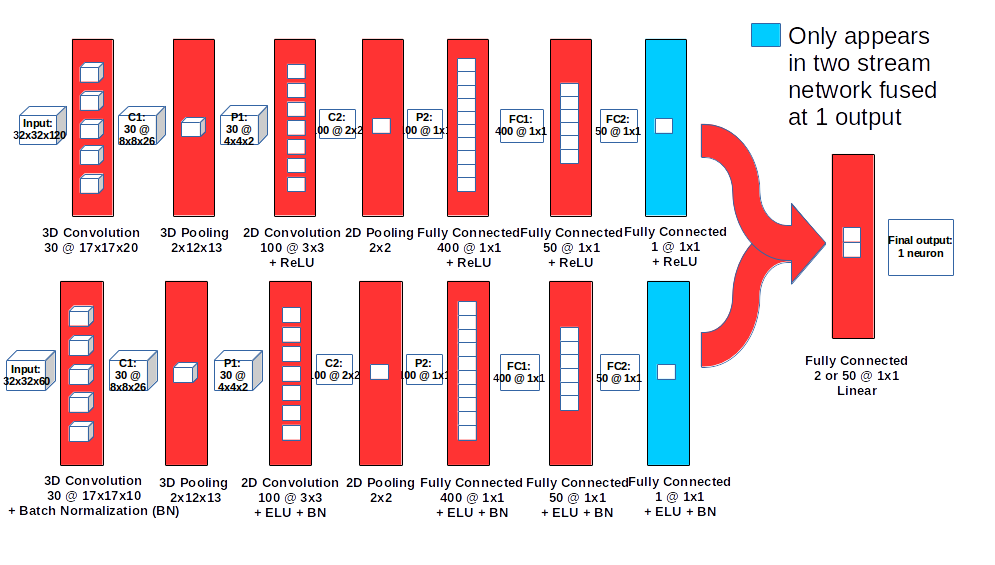}
	\centering
	\caption{Model of our two stream network, connected at 50 neurons or 1 neuron (layer added in blue), concatenated, and fully connected to a one neuron output.}
	\label{fig:twostream}
\end{figure*}

\section{Results}

\subsection{SVM}
\noindent Multiple trials with the SVM were conducted across different bin sizes and training sets.  While we compared the results using every 15th frame for static and dynamic backgrounds, as in Figure \ref{fig:svmresults}, our main interest was finding the feature descriptor and kernel that were most successful in classifying the approximate height of the camera.  Using a static background, Histograms of Oriented Gradient (HOG) with a linear kernel was most accurate in classifying the approximate height of a camera across all bin sizes. For 3 bins, 5 bins, and 11 bins, the HOG with a linear kernel classified with 64.69\%, 61.15\%, and 44.2\% accuracies, respectively.  See Figure \ref{fig:svmresults} for complete results.  This suggests to us that the angles of edges in the field of vision influence height estimation.  This makes sense because at different heights, the world is seen from different angles.  However, compared to the last layer of our spatial network and compared to the last layer of our spatial network adjusted to be a classifier, HOG did not do as well.  Because of this, we assume our networks are sufficient for predicting continuous height of the camera.

\begin{figure*}
\centering
\begin{subfigure}{.25\textwidth}
  \includegraphics[width=\linewidth]{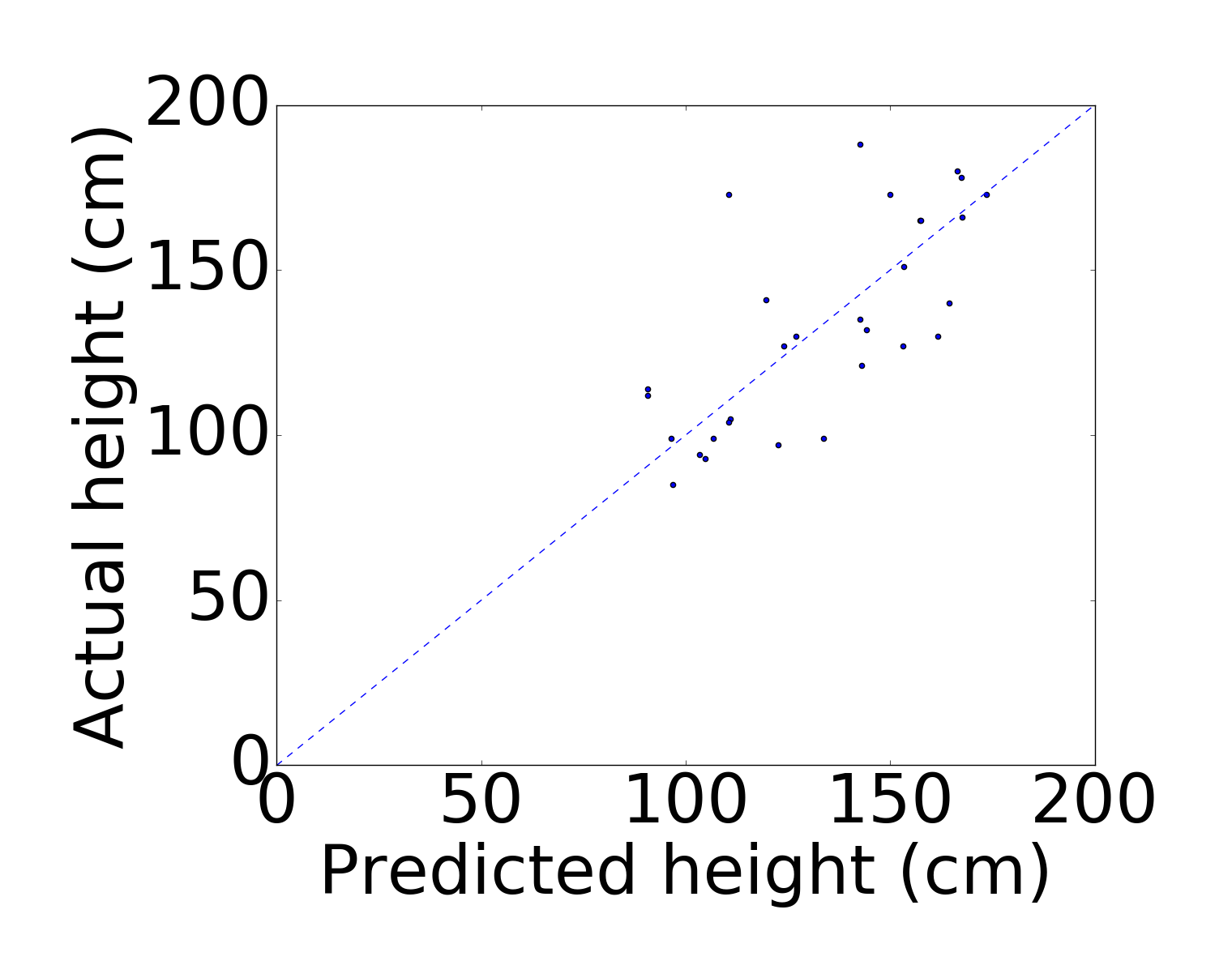}
	\caption{Temporal network}
	\label{fig:temporal}
\end{subfigure}\hfill%
\begin{subfigure}{.25\textwidth}
  	\includegraphics[width=\linewidth]{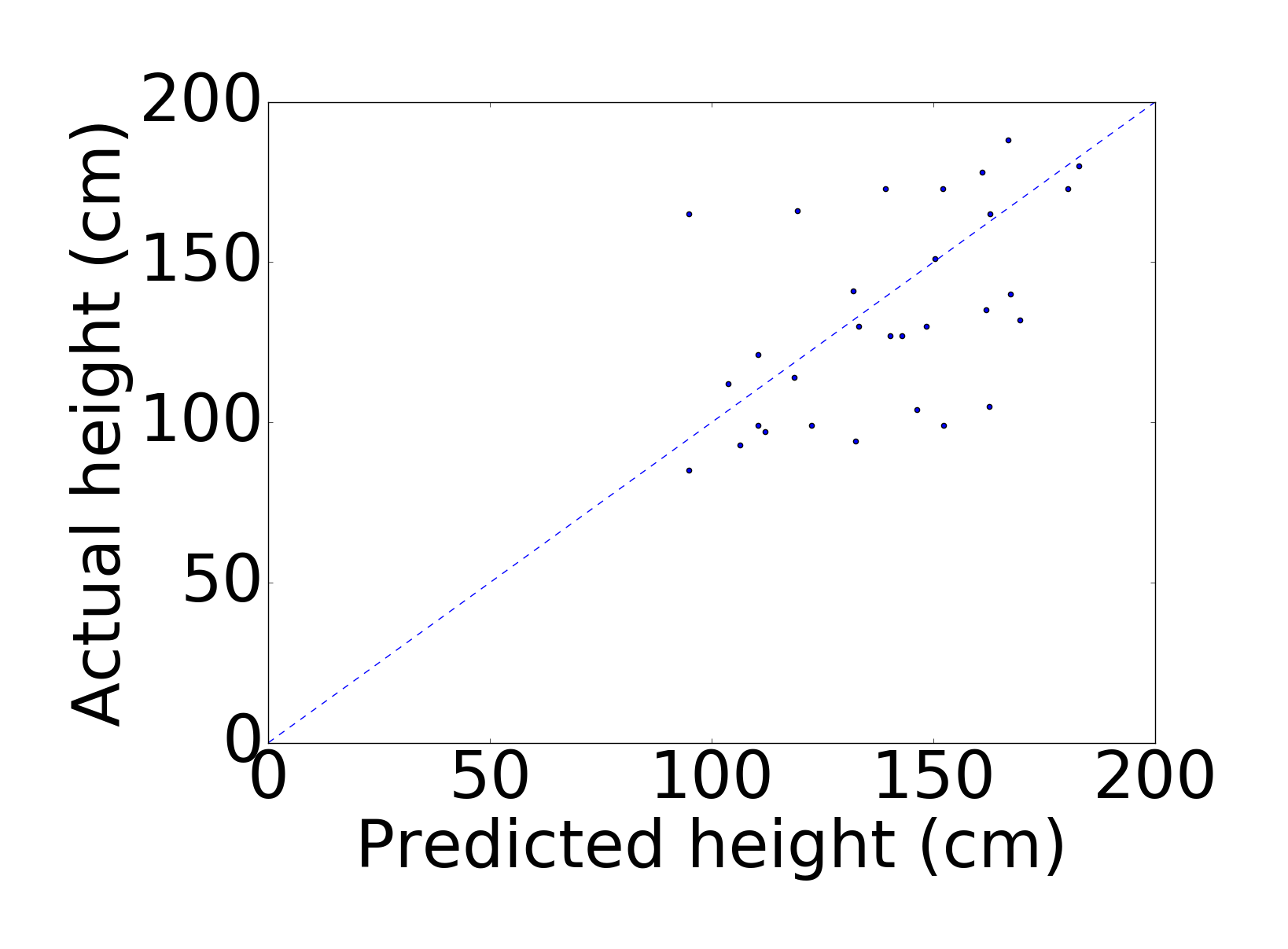}
	\caption{Spatial network.}
	\label{fig:spatial}
\end{subfigure}\hfill%
\begin{subfigure}{.25\textwidth}
  \includegraphics[width=\linewidth]{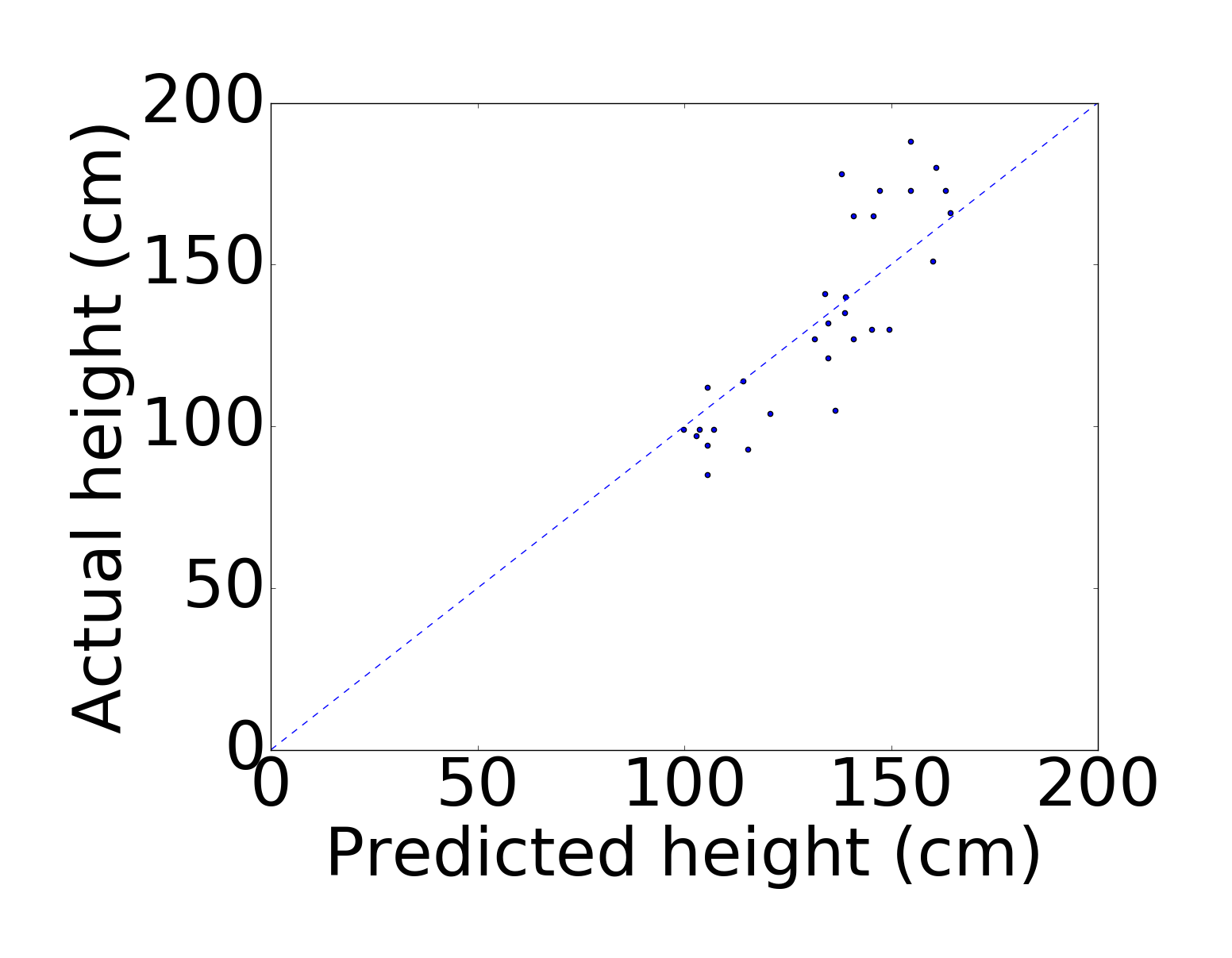}
	\caption{Two stream network connected at final output.}
	\label{fig:twostream1}
\end{subfigure}\hfill
\begin{subfigure}{.25\textwidth}
  \includegraphics[width=\linewidth]{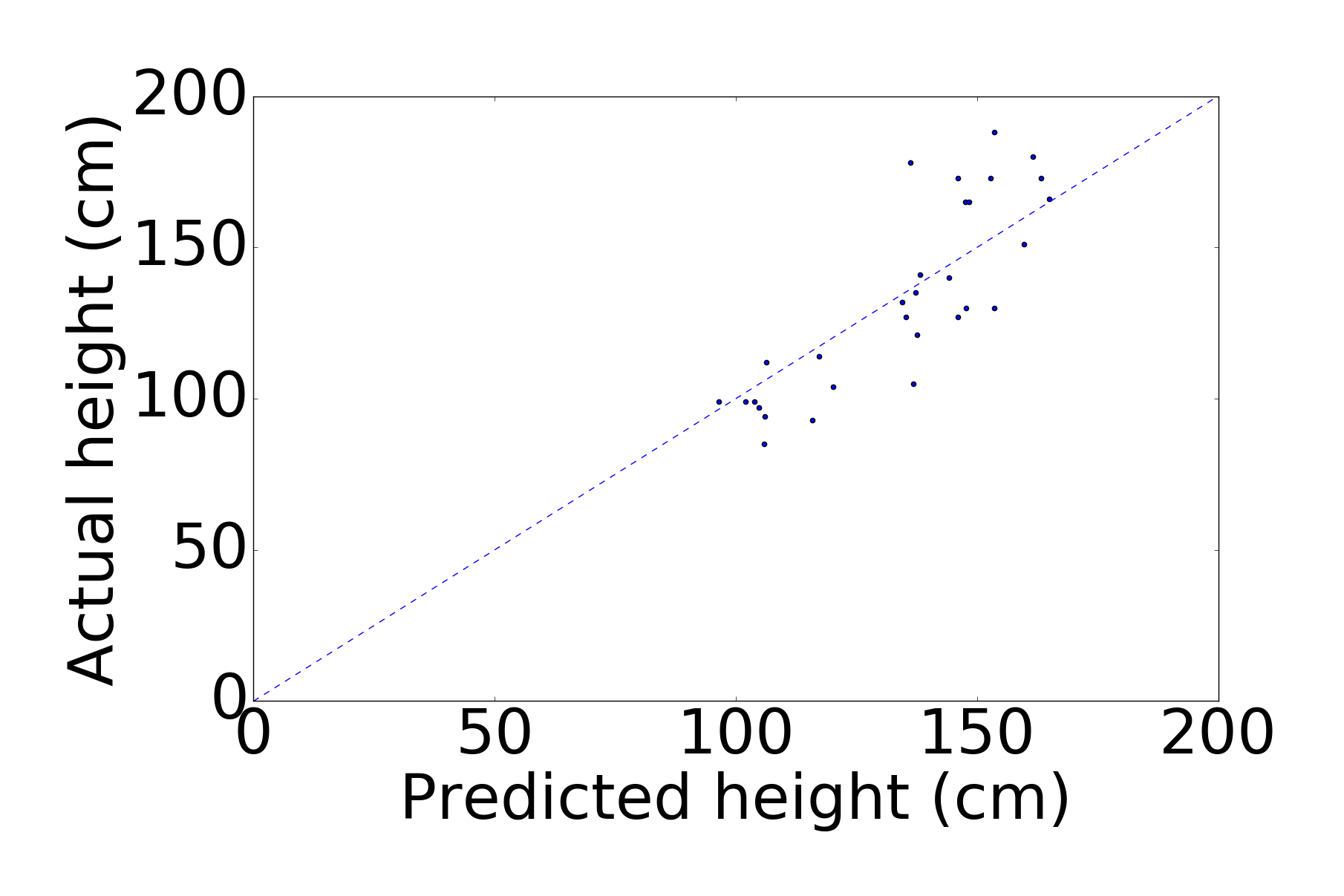}
	\caption{Two stream network connected at 50 neurons.}
	\label{fig:twostream50}
\end{subfigure}\hfill
\caption{LOO testing results (estimate v. actual) on different CNNs}

\label{fig:cnnresults}
	
\end{figure*}

\subsection{Neural Networks}
\noindent For each network, we performed Leave-One-Out (LOO) training by person on the dataset.  Note that for each round of LOO training, there are actually three videos being tested since we test by person.  There are usually 10-25 normalized video clips per video, and one estimate is yielded per video.  The estimate for the entire video is given by the average of the estimates for each clip.

\subsubsection{Temporal Network}
\noindent Using sparse optical flow as network input with LOO testing, the Mean Absolute Error (MAE) is 18.03 cm and Mean Squared Error (MSE) of 549.28.  The $r^{2}$ value is 0.4557.  A scatter plot of the estimates is shown in Figure \ref{fig:temporal}.

\subsubsection{Spatial Network}
\noindent With pixel intensities as network input, LOO testing yields a MAE value of 19.51 cm and MSE value 602.54.  The Temporal Network alone, therefore outperforms the Spatial Network.  However, a two-stream network outperforms either network by themselves.  $r^{2}$ with the spatial network is 0.4267.  The scatter plot of estimates is shown in \ref{fig:spatial}.

\subsubsection{Two stream network}
\noindent The two-stream network had LOO training performed twice.  Once, the network was only trained on the final outputs of each layer.  In this case, the MAE was 14.53 cm with a MSE value of 323.98 and $r^{2}$ value of 0.5773.  By all three metrics, the two-stream network outperformed both the spatial and temporal networks.  The scatter plot comparing estimated and actual heights for the two stream networks can be seen in \ref{fig:twostream50} and \ref{fig:twostream1}.\\

\noindent We also performed LOO training on the two-stream network, but instead of concatenating the outputs of each network, we trained it the second time based on the outputs of the fully connected layer that has 50 neurons to see if this improved results.  Here, we had MAE of 14.04 cm, MSE of 303.84, and an $r^{2}$ value of 0.6505.  These results all improve on the network when merged on the final output, albeit only slightly when considering MAE.\\

\noindent A full comparison of results can be seen in Table \ref{tab:compareresults}.

\begin{table}[]
\centering

\begin{tabular}{|l|l|l|l|}
\hline
           & MAE   & MSE    & $r^{2}$ \\
\hline
Temporal   & 18.03 & 549.28 & 0.4557  \\
\hline
Spatial    & 19.51 & 602.54 & 0.4267   \\
\hline
Two stream- 1 & 14.53 & 323.98 & 0.5773   \\
\hline
Two stream- 50 & \textbf{14.04} & \textbf{303.48} & \textbf{0.6505} \\
\hline
\end{tabular}
\caption{Comparison of LOO training results}
\label{tab:compareresults}
\end{table}

\section{Robustness}
\begin{figure*}
\centering
\begin{subfigure}{.25\linewidth}
  \includegraphics[width=\linewidth]{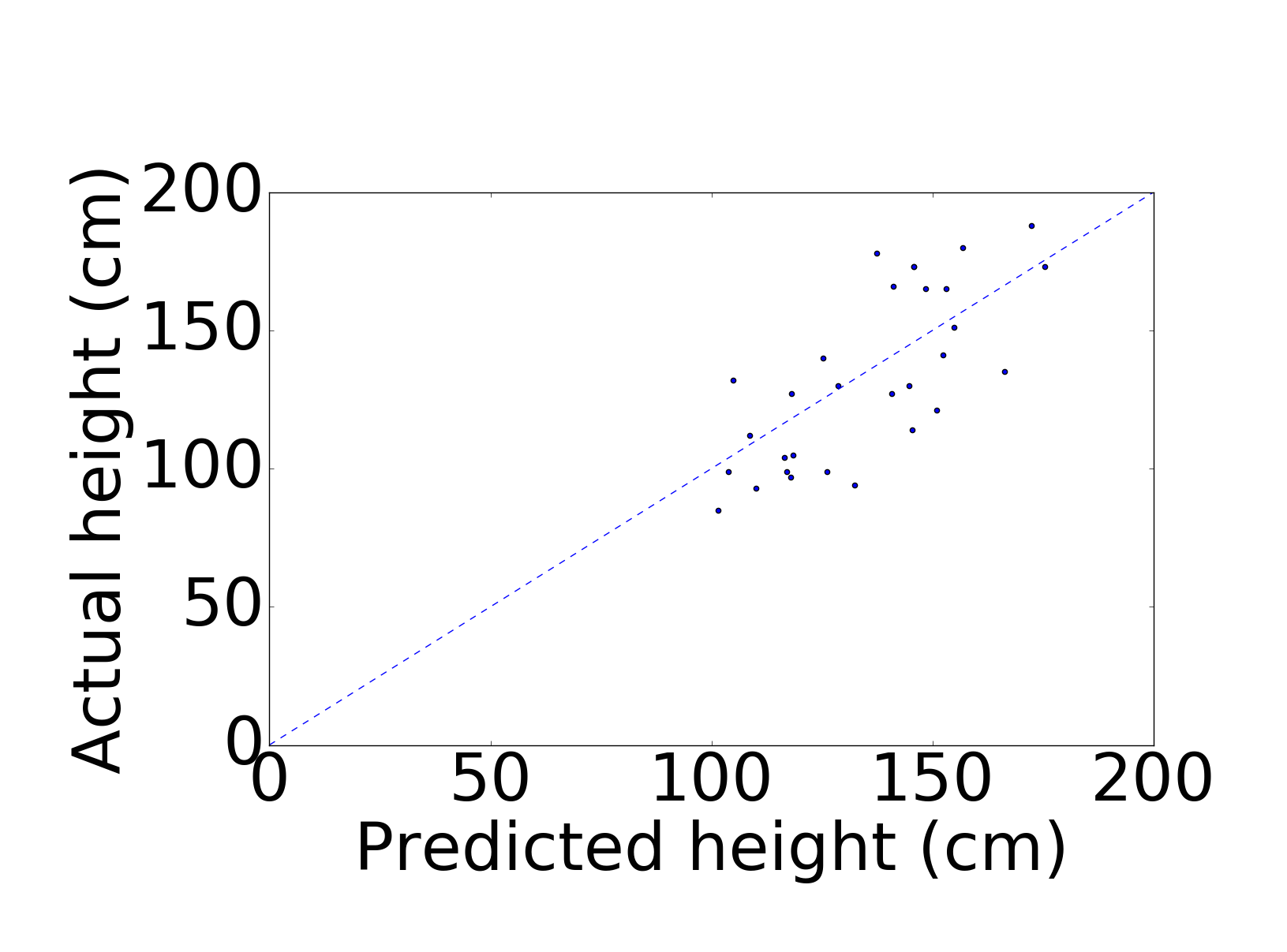}
	\caption{Temporal network trained of static data}
	\label{fig:temporalstatic}
\end{subfigure}\hfill%
\begin{subfigure}{.25\linewidth}
  	\includegraphics[width=\linewidth]{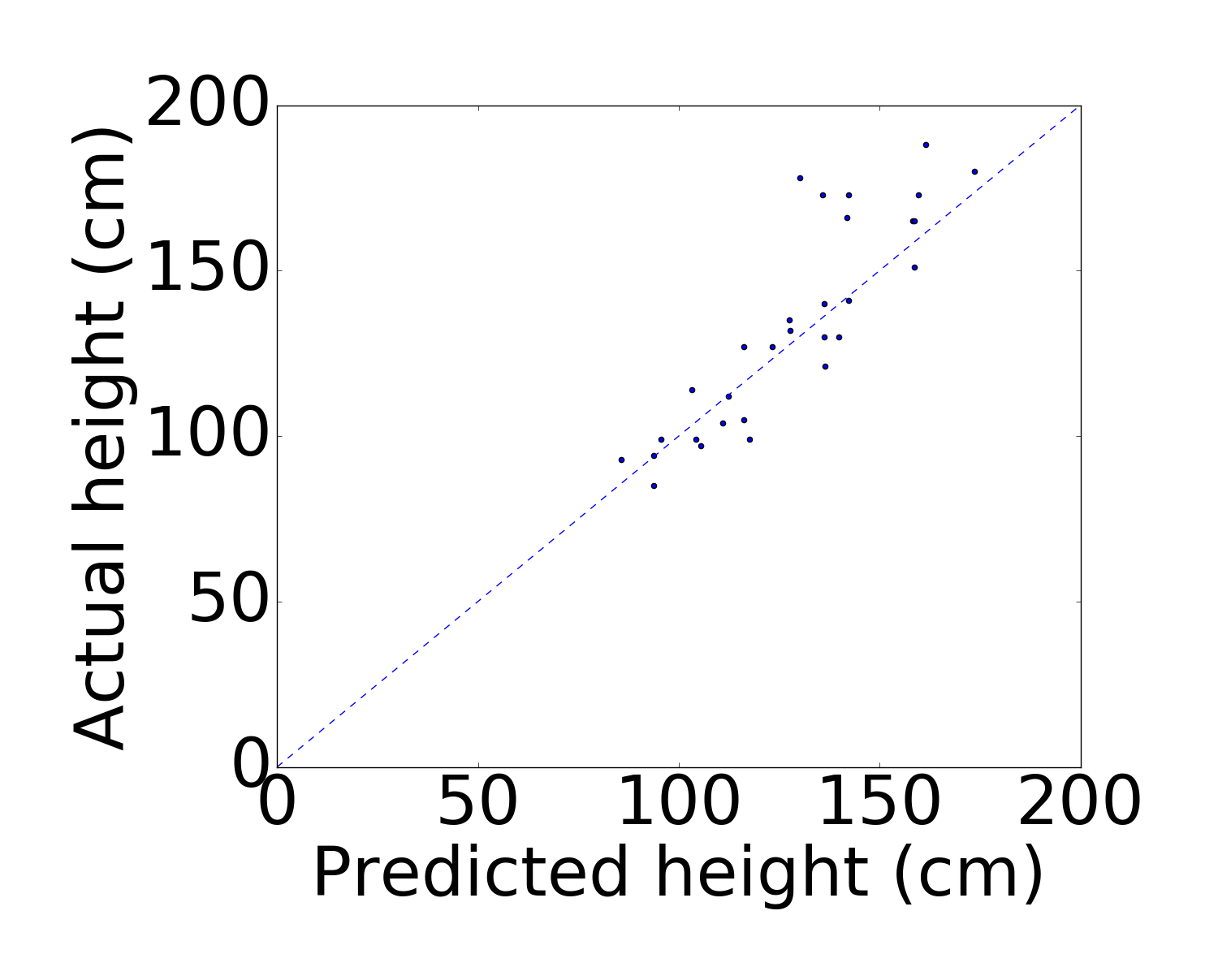}
	\caption{Temporal network trained on dynamic data.}
	\label{fig:temporaldyn}
\end{subfigure}\hfill%
\begin{subfigure}{.25\linewidth}
  \includegraphics[width=\linewidth]{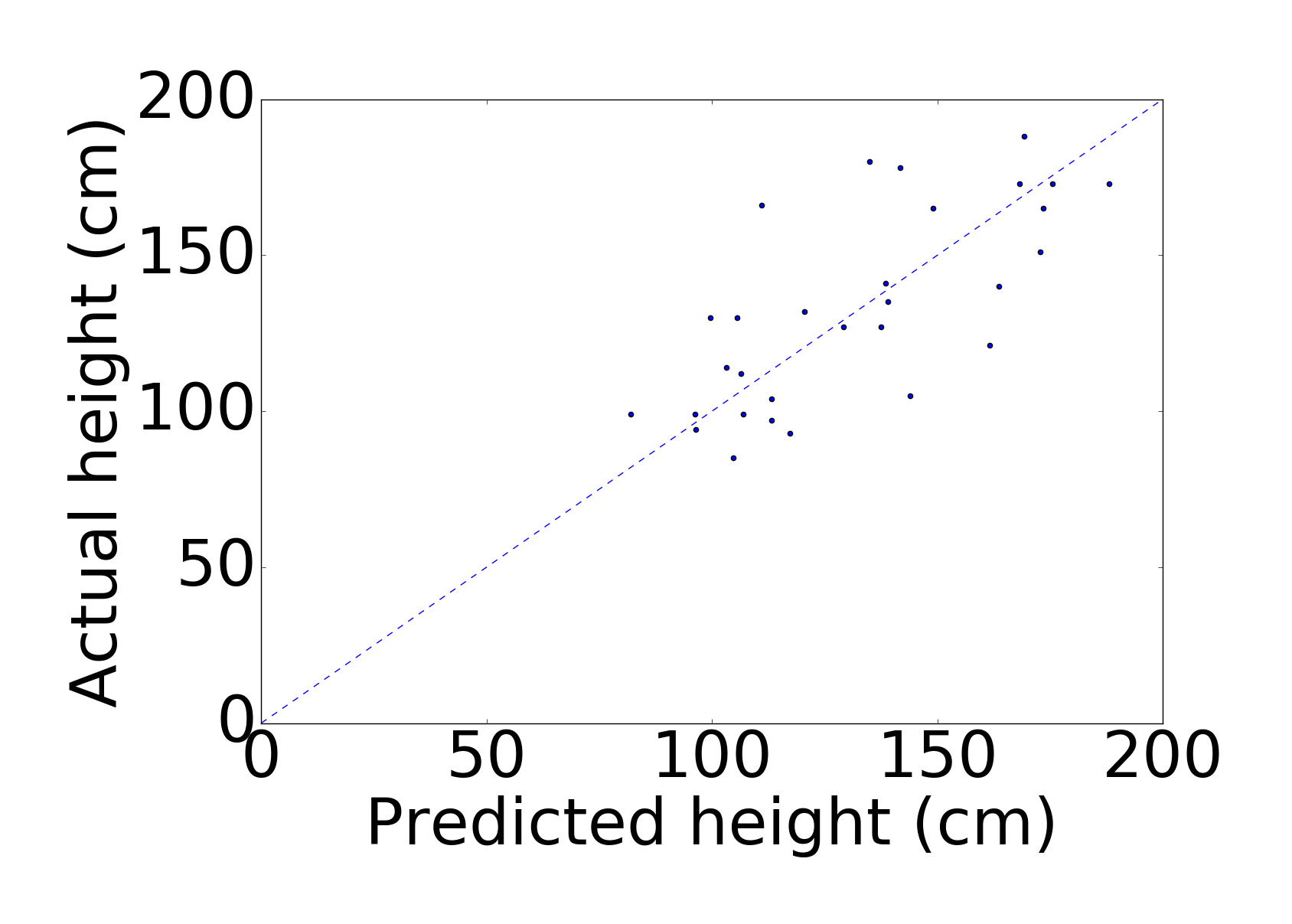}
	\caption{Spatial network trained on static data.}
	\label{fig:spatialstatic}
\end{subfigure}\hfill
\begin{subfigure}{.25\linewidth}
  \includegraphics[width=\linewidth]{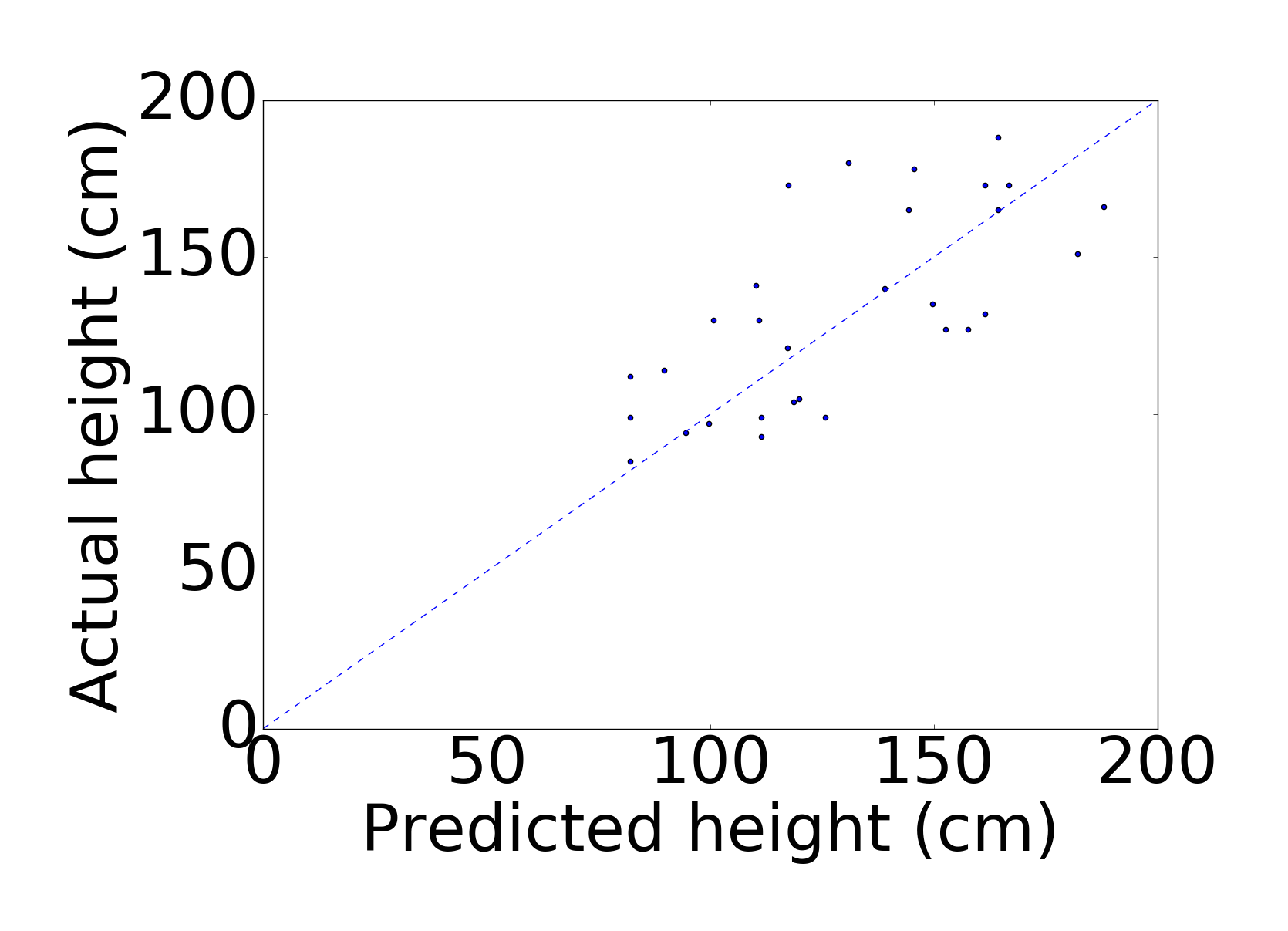}
	\caption{Spatial network trained on dynamic data.}
	\label{fig:spatialdyn}
\end{subfigure}\hfill
\caption{Training spatial and temporal networks on static and dynamic recordings.}

\label{fig:robustresults}
	
\end{figure*}

\noindent To test the adaptability of our networks, we took the two sets of data (one static and one dynamic) and estimated the height using our networks, training on one set and testing on the other.  As expected, MAE increased, but the amount of increase depends on the network and the training set.  Of these sets, the temporal set trained on dynamic data yielded the best results, with a MAE of 18.831 centimeters, close to our Leave-One-Out results on the temporal network that had a Mean Average Error of 18.03 cm.  However, training on static data and testing on dynamic data did not perform as well. Experimental results can be seen in Table \ref{tab:robustness} and are visualized in Figure \ref{fig:robustresults}.

\begin{table}[]

\begin{tabular}{|l|l|l|l|l|}
\hline
           & Train on  & MAE   & MSE    & $r^{2}$   \\
           \hline
Temp.   &Static & 27.403 & 1206.8268 & 0.5434  \\
\hline
Temp.   &Dynamic & \textbf{18.831} & \textbf{630.1156} & \textbf{0.7731}  \\
\hline
Spatial    &Static& 40.4695 & 2427.0375 & 0.5163 \\
\hline
Spatial    &Dynamic& 27.40375 & 1206.82684 & 0.4847  \\
\hline

\end{tabular}
\caption{Comparison of results training on different sets of dataset e.g. Row 1 shows results when trained on static and tested on dynamic background data.}

\label{tab:robustness}
\end{table}

\section{Conclusion and Future Work}
\noindent In this paper, we were able to extract important information regarding appearance of a person from his wearable camera.  While it is obvious that people see the world differently at different heights, we were able to train an SVM on different features to learn how a computer successfully perceives these differences in estimating height.  Additionally, we were able to train Convolutional Neural Networks based on temporal features, spatial features, and a two stream network which uses a combination of both spatial and temporal features to learn height estimation in egocentric video. Finally, we also present the first egocentric dataset with height labels in static and dynamic background. To our knowledge, this is the first research done on height estimation in egocentric videos and can be used to extract biometric information about a camera wearer. This further supports Peleg's claim in ~\cite{Peleg2} that egocentric camera wearers should share their videos with caution, because they might not as anonymous as they seem, despite rarely entering the frame.

\noindent  In future work, we plan to investigate different areas of biometric information extraction from egocentric video.  We want to investigate gait signatures, similar to the work of Peleg et. al. in ~\cite{Peleg2}.  Work could be done here to diagnose different gait disorders for people who may not be able to leave their homes with ease.  We are also interested in extracting the speed of an egocentric camera wearer in order to gain more biometric information and determine if a person was running, walking, or jogging.  Another interesting problem would be to estimate height of an egocentric camera based on detected vanishing points by exploiting the pinhole model of vision instead of using a convolutional neural network.

{\small
\bibliographystyle{ieee}
\bibliography{references}
}

\end{document}